\documentclass[letterpaper, 10 pt, conference]{ieeeconf}  

\IEEEoverridecommandlockouts                              

\usepackage{lipsum}
\usepackage{graphicx} 
\usepackage{epsfig} 
\usepackage{mathptmx} 
\usepackage{times} 
\usepackage{amsmath,amssymb,amsfonts}
\usepackage{cite}
\usepackage{pgfplots}
\pgfplotsset{compat=1.7}
\usepackage{pgfplotstable}
\usepackage{tikz,pgfplots}
\usepackage{physics} 
\usepackage{siunitx} 
\usepackage{enumerate} 
\let\labelindent\relax
\usepackage{enumitem}
\usepackage{tikz,pgfplots}
\usepackage{verbatim}
\usepackage{color}
\usepackage{mathtools}
\usepackage{breqn}
\usepackage{subcaption}
\usepackage{array}
\usepackage{amsmath}
\usepackage{makecell}
\usepackage{soul}  
\usepackage{comment}
\usepackage{textcomp}
\usepackage{subcaption}
\usepackage{algorithm}
\usepackage{algpseudocode}
\usepackage[labelsep=period]{caption}
\usepackage{setspace}
\usepackage{tabularx}
\usepackage[normalem]{ulem}

\usepackage[]{graphicx}
\usepackage{balance}

\usepackage{kotex} 

\newcommand{\sg}[1]{\textcolor{black}{#1}} 

\title{\LARGE \bf
EIT–Pneumatic Hybrid Robotic Skin for Practical and Accurate\\ Force Map Reconstruction}

\author{$^{\dagger}$Junhwi Cho, $^{\dagger}$Sunggyu Bae, $^{\dagger}$Junghyeon Ma, Hyosang Lee, Jung Kim, and $^{*}$Kyungseo Park
\thanks{J. Cho and J. Kim are with the Mechanical Engineering Department, KAIST (Korea Advanced Institute of Science and Technology), 34141 Daejeon, Republic of Korea}
\thanks{S. Bae, J. Ma, and K. Park are with the Department of Robotics and Mechatronics Engineering, DGIST (Daegu Gyeongbuk Institute of Science and Technology), 42988 Daegu, Republic of Korea}
\thanks{H. Lee is with the Mechanical Engineering Department, TU/e (Eindhoven Technical University, 5600 MB Eindhoven, the Netherlands}
\thanks{$^{\dagger}$These authors constributed equally to this work.}
\thanks{$^{*}$Corresponding authors: kspark@dgist.ac.kr}}

\begin{document}
    \maketitle
    \thispagestyle{empty}
    \pagestyle{empty}
    
    \begin{abstract}

We present a hybrid robotic skin that combines electrical impedance tomography (EIT) with pneumatic tactile sensing to improve force reconstruction capability. The developed robotic skin is fabricated entirely by 3D printing and spray coating, making it affordable and easy to build. A Tikhonov-regularized inverse reconstruction, paired with per-pad pneumatic calibration, enables accurate large-area tactile sensing with a simple measurement scheme. For validation, we conducted load-cell indentation experiments; the results showed consistent force reconstruction across locations within a pad. \sg{Compared with an EIT-only baseline,} sensitivity non-uniformity was also reduced, with the coefficient of variation decreasing from 0.31 to 0.14, indicating that the proposed approach addresses a longstanding limitation of EIT. We further demonstrated chest-mounted integration on a humanoid robot and found that the pneumatic signals remained reliable across diverse contact scenarios, including multiple simultaneous contacts on the same sensing pad. These results indicate a practical path toward accurate, scalable whole-body tactile sensing in real robotic systems.


\end{abstract}
    
    \section{INTRODUCTION} 

    Physical human–robot interaction (pHRI) is an important factor in deploying robots in everyday environments. For safe and dependable interaction, robots must not only perceive physical contact without blind spot but also make the contact itself gentle and pleasant through material and structural design \cite{Haddadin2011Safe, park2011designing, svarny2022effect}. This motivates the need for practical whole-body robotic skin, which provides continuous tactile perception across large areas and protective function simultaneously, rather than focusing solely on tactile sensing.

    Among various tactile sensing methods, electrical impedance tomography (EIT) has been a promising method\cite{tawil2015electrical, liu2020artificial, cui2023recent}. EIT-based robot skin uses a continuous piezoresistive surface in which distributed electrodes inject current and measure voltage \cite{costa2023variable, chen2023novel}. The conductivity distribution can then be reconstructed from measurements using optimization\cite{vauhkonen1998tikhonov,borsic2010invivo,liu2018limage} or deep neural network\cite{li2019novel, park2021Deep}. It allows electrodes to be arranged freely, making EIT suitable to large and curved geometries\cite{tawil2011improved, nagakubo2007deformable }. Moreover, fragile sensing elements can be physically isolated from external stimuli, enabling a durable structure and a clean exterior \cite{kim2024extremely}. However, strong sensor nonlinearities complicate signal processing \cite{hrabuska2018image, duan2019artificial}, and fabrication still requires considerable manual effort \cite{park2020ERT}. 

    Meanwhile, pneumatic robot skin is also a highly practical solution. It employs an air-tight, compliant structure with internal cavities and monitors changes in internal air pressure under external forces \cite{Kim20153D,goncalves2022punyo}. The skin is lightweight and inherently shock-absorbing, which are keys for safe physical human-robot interaction\cite{rustler2024adaptive}. It is also inexpensive, easy to fabricate, and conceptually simple. Combined with 3D printing, it supports multi-material, customization and automated production\cite{park2024lowcost}. These advantages align well with our goals for practical and deployable systems. Nonetheless, it cannot localize contact within a single pad, resulting in limited spatial resolution.

    \begin{figure*}[htbp]
        \centering
        \includegraphics[width=6.9in]{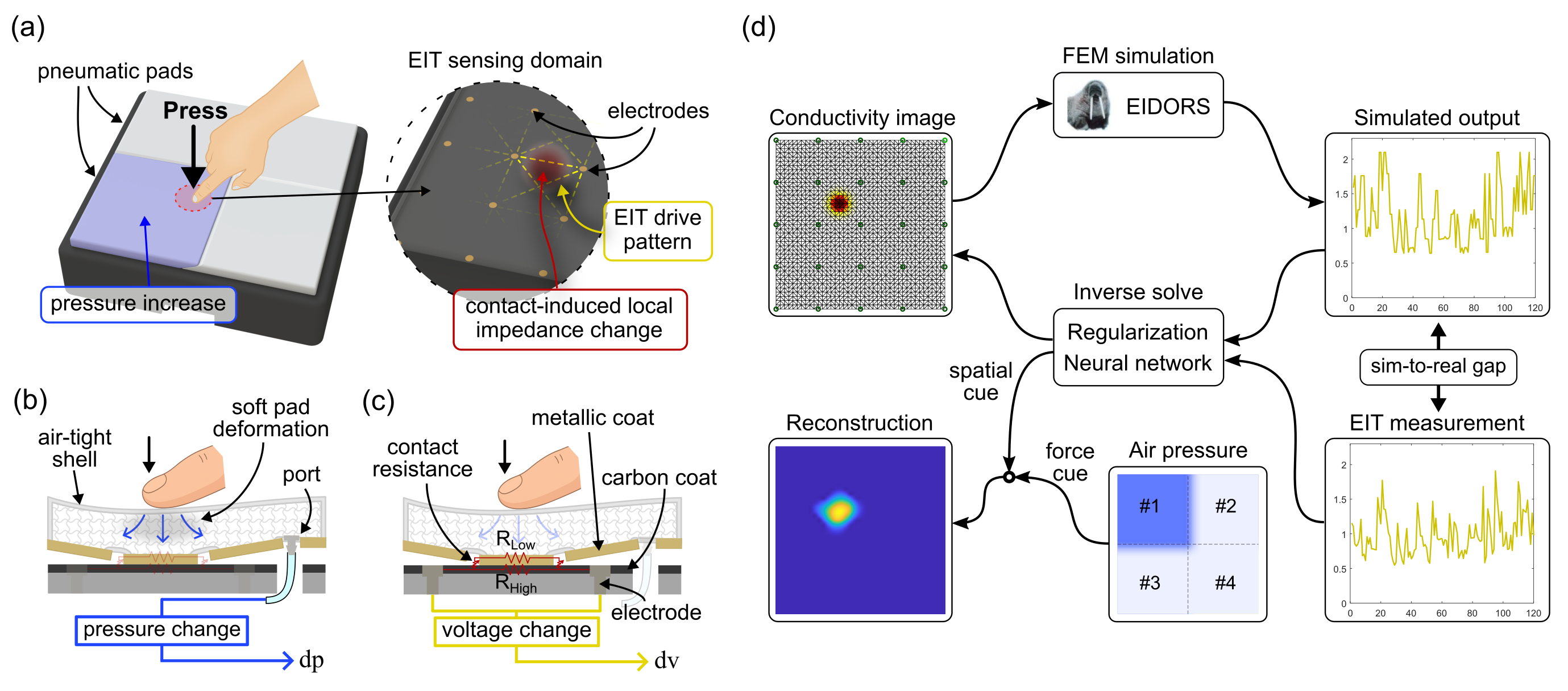}
        \caption{Inferential tactile sensing concept. (a) EIT-pneumatic hybrids composed of rigid base and soft pad; (b) pneumatic tactile sensing utilizing a soft, air-tight pad connected with an air-pressure sensor; (c) piezoresistive tactile sensing (EIT) based on contact-resistance; (d) Multi-modal sensor fusion pipeline using Tikhonov-regularized reconstruction with pad-wise calibration\sg{.}}
        \label{figure label 07}
    \end{figure*}

    To address these limitations, we propose a hybrid tactile sensing system that combines electrical impedance tomography (EIT) and pneumatic modalities for practical and accurate force reconstruction. The two modalities are complementary: EIT offers a relatively fine spatial cues, while the pneumatic pad provides accurate and sensitive force measurements. Prior work has explored combining EIT- or pneumatic-based tactile sensors with other modalities, demonstrating benefits such as improved gesture recognition and interaction capabilities.\cite{park2022biomimetic, park2024fully, yang2025body, chen2025large}.
    
    Notably, our robotic skin was fabricated entirely by digital manufacturing, yielding an accessible and practical solution suitable for integration in real robotic systems. Moreover, this paper proposes a sensor fusion method, which utilizes Tikhonov-regularized reconstruction from EIT and pad-wise calibration using pneumatic sensing. As this approach does not require complex modeling or extensive data collection, it is very straightforward and accessible to various robotics engineers, while its contact force estimation is more accurate than in uni-modal cases.

    The remainder of this paper is organized as follows. Section II presents the proposed inferential tactile-sensing framework and its EIT–pneumatic fusion scheme. Section III describes the hardware design and fabrication process. Section IV presents experimental results, and Section V demonstrates tactile sensing capabilities in diverse settings. Finally, Section VI concludes with a discussion and directions for future work.

    \section{INFERENTIAL TACTILE SENSING}

    \subsection{EIT--Pneumatic Hybrid}
        EIT-based robot skin reconstructs the spatial distribution of electrical conductivity within a continuous piezoresistive domain by observing boundary voltages under actively injected currents. Since each measurement inherently covers an overlapping receptive field, the spatial distribution of contact-induced resistance changes can be inferred numerically. This inference requires sufficient information of the mapping from conductivity to measured signals, which is impractical to obtain experimentally. Therefore, many researchers have relied on a forward model, typically realized by a finite element (FE) model \cite{adler2006eidors}.

        \begin{equation}
            \Delta \mathrm{V} = f(\Delta \sigma),
        \end{equation}
        
        where $\Delta \sigma$ and $\Delta \mathrm{V}$ denote the changes in conductivity distribution and the resulting boundary-voltage difference, respectively. Once simulation data is obtained, we solve the inverse problem to reconstruct $\Delta \sigma$ from observed $\Delta \mathrm{V}$, either through optimization-based solvers \cite{vauhkonen1998tikhonov,borsic2010invivo,liu2018limage} or machine learning frameworks, such as deep neural networks (DNNs) \cite{park2021Deep, zheng2025large, dong2025learning}.

        From an AI perspective, EIT reconstruction can be framed as learning a latent representation from complex measurements and mapping it to interpretable physical quantities. When the forward model is well characterized, large labeled datasets can be synthesized via FE model, enabling simulation-based training and sim-to-real transfer. This enables \textit{tactile super-resolution}, in which high-resolution contact images are inferred from limited measurements. However, EIT suffers from several limitations, including nonlinear piezoresistive behavior, spatial inhomogeneity within the conductive domain, and the high computational cost of dense measurements and reconstruction. These issues are a nontrivial problem in a practical sense.   
        
        To mitigate these limitations, we propose to integrate EIT with pneumatic tactile sensing method. Pneumatic signals provide global force cues with high sensitivity, wide dynamic range, and improved accuracy, although it cannot localize contact location. However, a reliable estimation of the contact force of the pneumatic sensor can complement the inferior one of the EIT. 

    \subsection{Data Sampling}
        
        Typically, EIT treats conductivity perturbations $\Delta \sigma$ as inputs and computes the corresponding boundary voltage differences $\Delta \mathrm{v}$ using finite element simulation. We generate data using EIDORS (MATLAB toolkit for EIT)\cite{adler2006eidors}, assuming a spatially uniform baseline conductivity $\sigma_{0}$ equal to the measured average conductivity of the carbon surface. The perturbation amplitude was from zero to $\sigma_{0}$ (i.e., up to 100 $\%$ of the baseline), following prior works. Each perturbation had a Gaussian spatial profile to approximate the sub-surface load spreading induced by the pneumatic pads. The perturbations were placed on a 200 mm $\times$ 200 mm planar mesh; centers were sampled every 5 mm in both axes, resulting in a 41 $\times$ 41 grid. The finite element mesh used in our experiments is shown in Fig.1(d).
       
        Compared to EIT, pneumatic signals are much easier to acquire in reality, thus they are not simulated at this stage. Since the pneumatic signal serves as a proxy for the applied force, we collected paired measurements by pressing the pneumatic pad with an indenter assembly, thereby recording the contact location, the synchronized pneumatic pressure, and the corresponding ground-truth force. During data collection, the indenter was mounted on a motorized stage and followed a predetermined trajectory on the surface of the pad. \sg{The experimental setup is described in Section IV-A.}
        

        



        \begin{figure}[t!]
            \centering
            \includegraphics[width=3.3in]{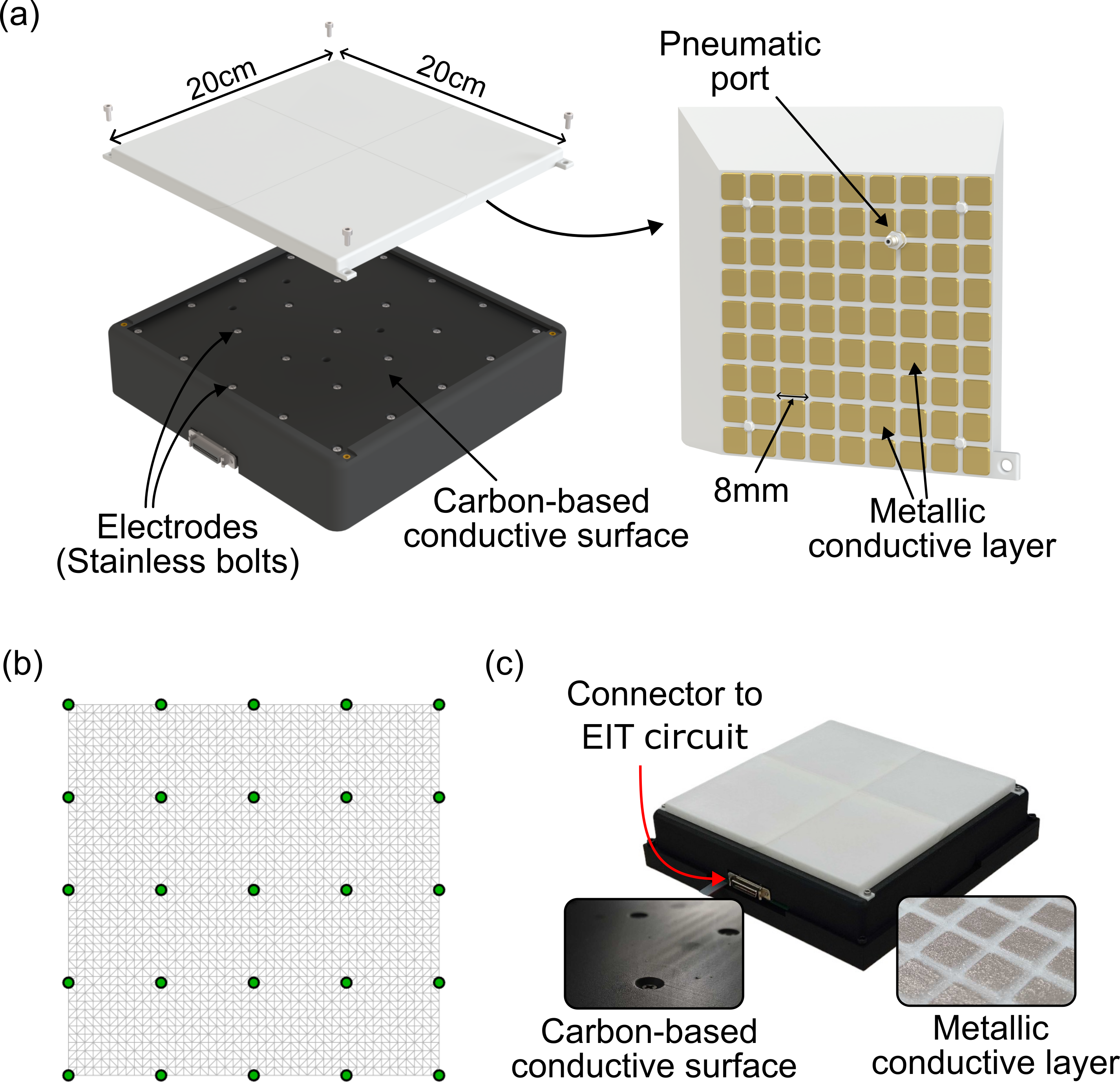}
            \caption{Structure of the EIT-pneumatic hybrid sensor. (a) rigid base layer and soft pneumatic pad; (b) FE mesh of EIT simulation model, (c) photograph of the assembled sensor system\sg{.}}
            \label{figure label 03}
        \end{figure}   
        
    \subsection{Reconstruction and Calibration}
        The goal of our sensor-fusion scheme is to leverage pneumatic signals to improve the accuracy of the force map estimation for EIT-based reconstruction. Classical EIT requires solving an inverse problem, and the regularization often introduces severe distortions to the result. In addition, fabrication of the EIT tactile sensor create a non-negligible sim-to-real gap, yielding artifact, non-uniform sensitivity, and excessive blurring that are difficult to remedy.

        In particular, EIT reconstructions provide the spatial distribution of the applied load but their absolute magnitude is not reliable. Although one can train a separate DNN to regress forces\cite{lee2020calibrating}, a more principled approach is to incorporate an additional sensing modality with a direct force measurement. A pneumatic pad offers more reliable contact force measurements over large areas.

        First, we solve the EIT inverse problem to obtain a conductivity-perturbation image:

        \begin{equation}
            \Delta \hat{\sigma} = Q\Delta v, \quad Q=(J^TJ+\lambda^2 \Gamma^T \Gamma)^{-1}J^T
        \end{equation}

        where $J$ is the Jacobian (sensitivity) matrix mapping conductivity perturbations and to boundary-voltage differences, and $Q$ is the Tikhonov-regularized reconstruction matrix with a hyperparameter $\lambda$ and prior $\Gamma$ (e.g., NOSER type) \cite{vauhkonen1998tikhonov}. 

        Concurrently, we read pneumatic signals from soft pads. The sensing area is partitioned into four pads; only the indented pad exhibits a pressure change. We assume the pneumatic change is proportional to the net force applied to that pad, since the pressure variation reflects a volume change in the air-tight shell.

        Let $M_{i}(r)\in[0,1]$ denote the spatial footprint (mask) of pad $i$, where $r=(x,y)$ denotes a location on the 2D sensor plane. Then, We form pad-specific weighted conductivity images
        \begin{equation}
            \Delta \hat{\sigma}_{i}(r) \;=\; M_{i}(r) \; \odot \;  \Delta \hat{\sigma}(r)
            \label{eq3}
        \end{equation}
        
        and distribute the estimated pad-level force $F_{i}$ over its footprint to obtain a calibrated force map
        \begin{equation}
            \hat{\sigma}_{adj}(r) = \sum_{r \in \Omega_{i}} F_{i}  \cdot  \frac{\Delta \hat{\sigma}_{i}(r)}{\sum{\Delta \hat{\sigma}}_{i}(r)}
            \label{eq4}
        \end{equation}

        This procedure confidently suppresses artifacts in regions where load is not applied and uses the pneumatic signals to calibrate the level of the EIT-derived image while preserving its spatial detail.



       

        
    \section{SENSOR HARDWARE}   

    \begin{figure}[t!]
        \centering
        \includegraphics[width=3.3in]{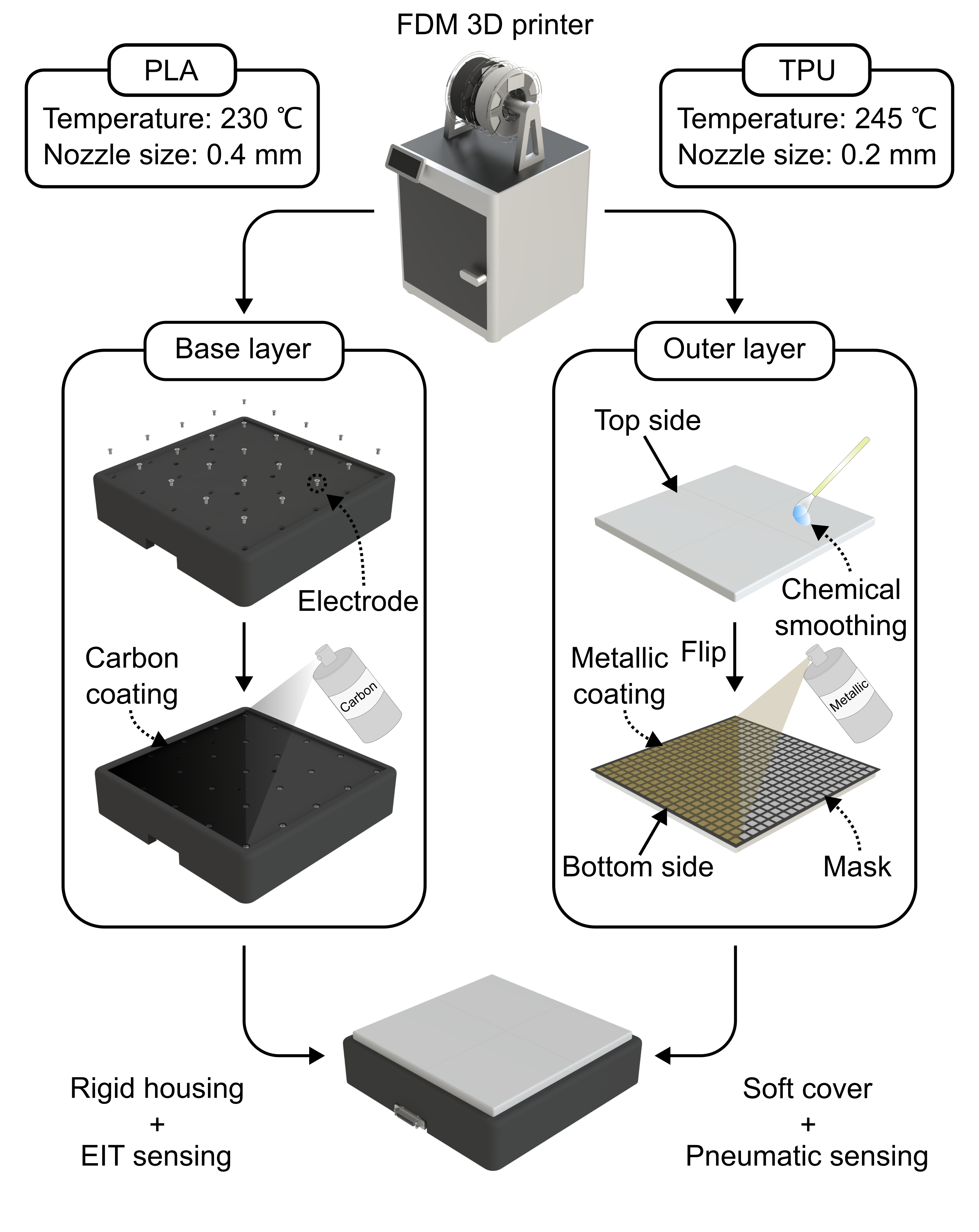}
        \caption{Fabrication of the EIT-pneumatic hybrid skin. The base and outer layer are printed in PLA and TPU, respectively. They are then spray-coated with carbon and metal ink to realize a contact-based piezoresistive structure.}  
        \label{figure label 04}
    \end{figure}
        
    A sensor testbed was designed and fabricated to validate the proposed sensor fusion framework; the rendered image, forward model (i.e., mesh), and prototype are shown in Fig.2. The prototype adopts a two-layer structure that allows simultaneous electrical impedance tomography (EIT) and pneumatic sensing. The base layer provides the conductive substrate and electrode interface for EIT, while the outer layer converts external forces into local contact resistance changes and also interfaces with pneumatic ports for direct pressure measurement. 
    
    The base layer has dimensions of $200 \times 200$~mm and is fabricated from a polylactic acid (PLA) filament using fused-deposition 3D printing. The PLA surface is spray-coated with a carbon-based conductive material (838AR Total Ground, MG chemical), forming a continuous conductive domain. A $5 \times 5$ array of stainless-steel bolts is installed as electrodes, each fixed with a nut and internally wired to the EIT driving circuit. Through an external multiplexer, arbitrary electrode pairs can be selected for current injection and voltage measurement.  

    \begin{figure}[t!]
        \centering
        \includegraphics[width=3.1in]{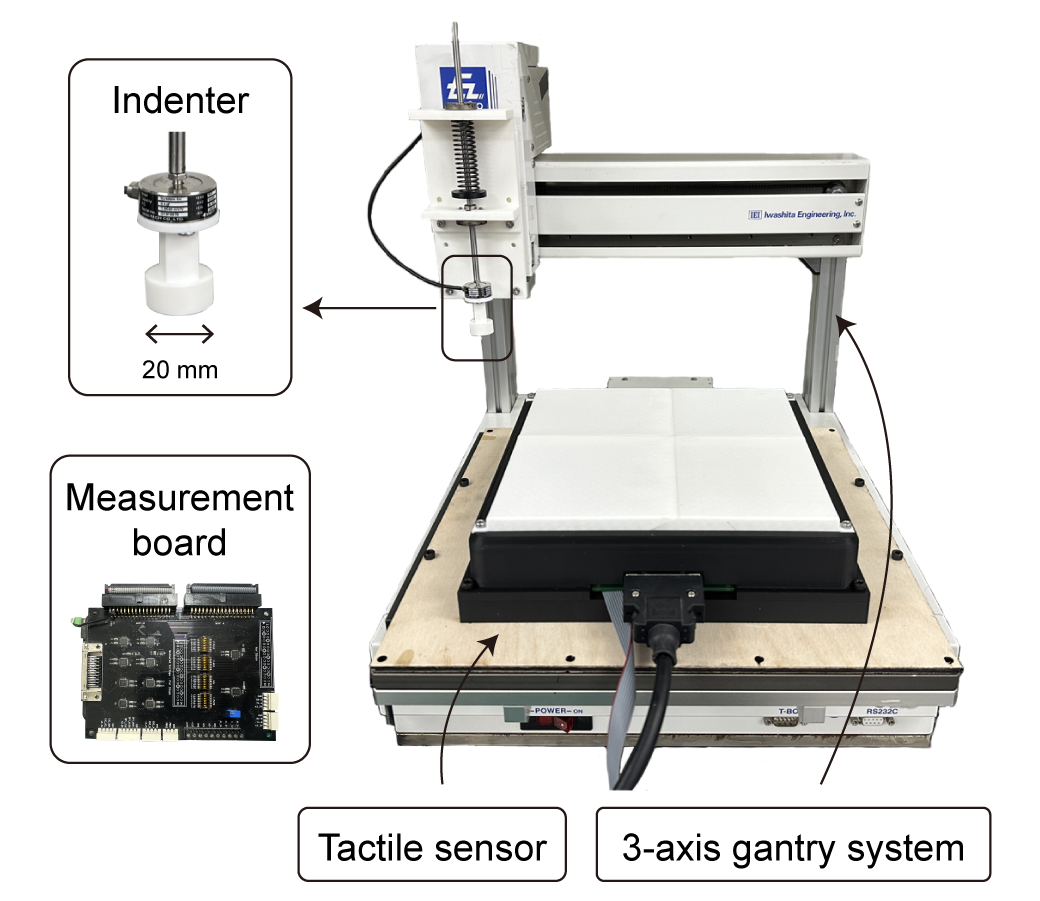}
        \caption{Indentation setup including the motorized stage and the indenter assembly. The testbed is placed on the stage and repeatedly indented along a predefined path.}
        \label{figure label 05}
    \end{figure}

    The outer layer has the same footprint ($200 \times 200$~mm) with a thickness of 1~cm and is fabricated from thermoplastic polyurethane (TPU). Its inner surface, facing the base layer, is patterned with a regular grid of square protrusions. These protrusions are spray-coated with a metallic conductive paint (843AR Super Shield, MG chemical), such that external forces generate the contact resistance between the base surface (carbon) and highly conductive surface (metallic). Consequently, current flows through this contact resistance into a highly conductive surface, resulting in a local conductivity change. 
    In parallel, embedded pneumatic ports connect to external pressure sensors through silicone tubing and barbed fittings, enabling direct measurement of air-pressure changes induced by contact.

    All structural components are produced by 3D printing, with PLA used for the rigid base and TPU for the compliant outer layer. The PLA substrate is sanded to improve surface uniformity, while the TPU layer is post-processed with an organic solvent for surface smoothing. Two types of conductive coatings are employed: carbon-based spray for the base surface and metallic spray for the highly conductive surface on protrusions. Electrodes are secured mechanically with bolts and nuts to ensure robust connections. Pneumatic channels are implemented using silicone tubing, and airtight connections are achieved with barbed ports. Pressure measurements are obtained using air-pressure sensors (HSCDANT005PGAA5, Honeywell). The overall fabrication process of the prototype is illustrated in Fig.3.

    \section{EXPERIMENTS AND RESULTS}
    \begin{figure}[tb!]
            \centering
            \includegraphics[width=3.1in]{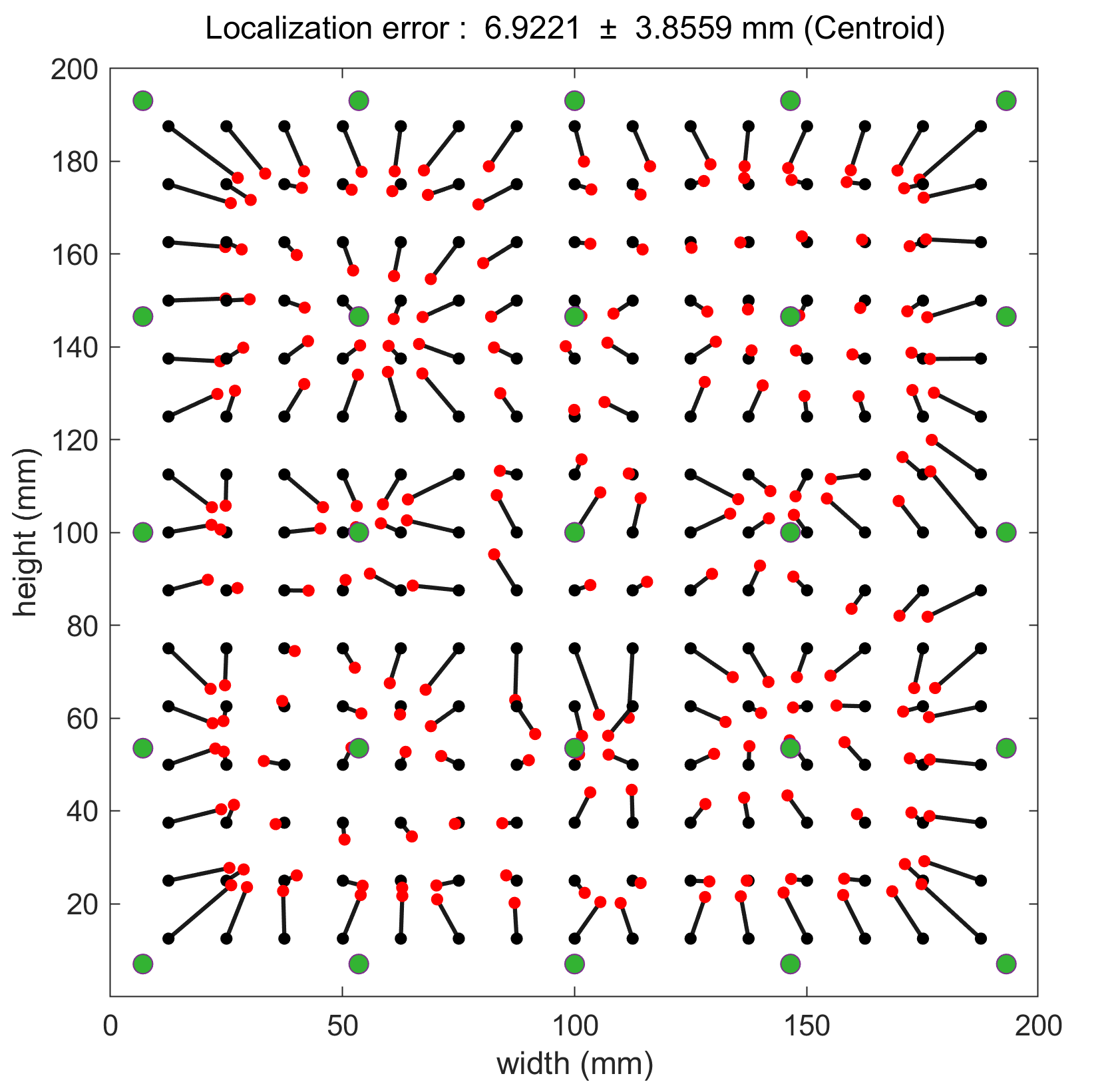}
            \caption{Single-point contact localization. The average error was 6.9 mm. The black and red dots indicate the true and estimated location of contacts. The green circles indicate the locations of electrodes arranged in 5-by-5 grid.}
            \label{figure label 05}
        \end{figure}
    \subsection{Experiment setup}

        
        
        

        To evaluate the proposed sensing framework, we conducted indentation experiments using a motorized linear stage and a custom indenter unit, as illustrated in Fig.4. The stage (EzRobo-5GX, Iwashita Engineering, Japan) provides precise three-axis positioning of the indenter over the sensor surface. The indenter unit consists of a cylindrical tip (diameter 20mm) mounted on a spring-guided fixture and integrated with a miniature load cell (capacity 50N, resolution 0.01N), enabling accurate measurement of normal force during contact.  
        
        During each test, the stage controlled the indenter position in a raster-scanning manner throughout the sensing area. At each indentation point, synchronized data streams were acquired: (i) EIT boundary voltage measurements via the multiplexer circuit, (ii) pneumatic pressure from the embedded air-pressure ports, (iii) ground-truth contact force from the load cell, and (iv) stage position and time stamps. The EIT system operated at a sampling period of 10µs per measurement, and pneumatic signals were digitized at the same frequency to allow synchronized fusion.  
        
        A standard indentation protocol was set to ensure repeatability. The cylindrical tip indented the sensor system up to 20N, was maintained for 2 seconds, and then released. The indentation was repeated on a $15 \times 15$ grid covering the entire $200 \times 200$~mm sensor surface, resulting in 225 single-point contact locations. Blank measurement is also collected to estimate noise characteristics. As post-signal processing, a low-pass filter with 50 Hz cut-off frequency was applied to the raw signals, and then baseline drift was compensated prior to reconstruction.  
        
        \begin{figure}[tb!]
            \centering
            \includegraphics[width=3.3in]{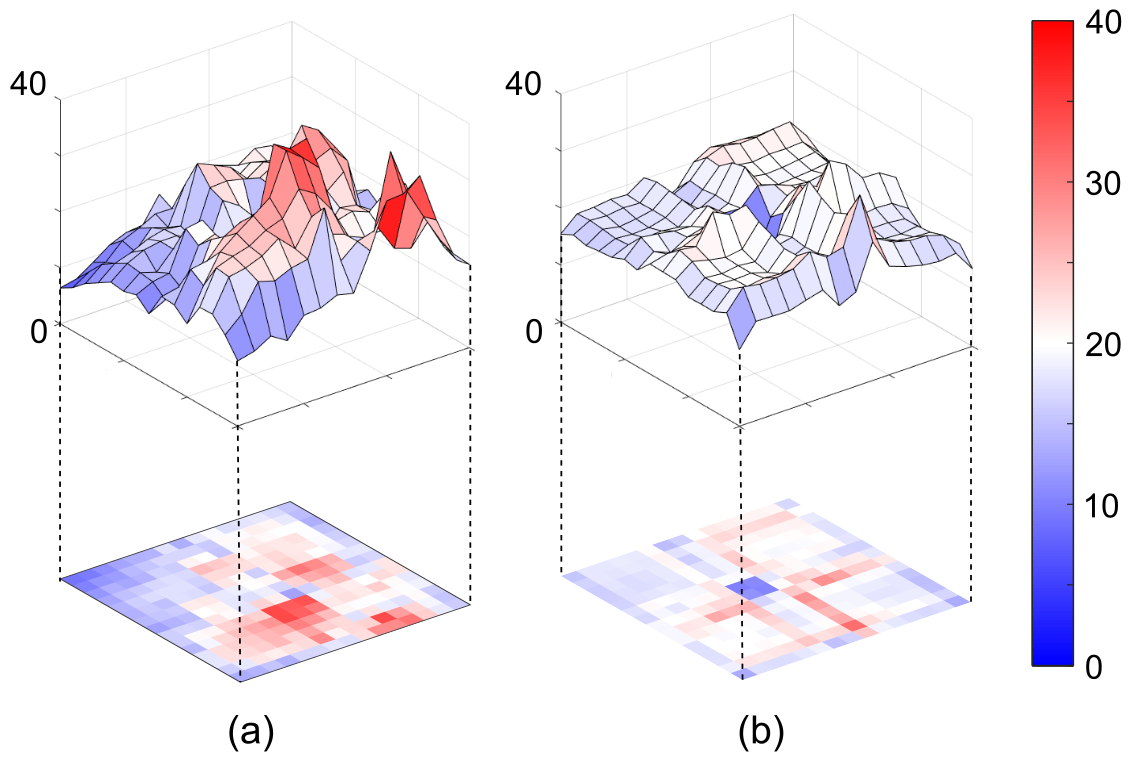}
            \caption{Reconstruction sensitivity map. (a) EIT-only result (CV = 0.31); (b) EIT-pneumatic hybrid (CV = 0.14)\sg{.}}
            \label{figure label 06}
        \end{figure}
        
    \subsection{Contact localization and force estimation}
        

        We conducted single-point contact experiment to evaluate the contact localization and force estimation performance. In particular, influence of the integration of a pneumatic sensor with an EIT-based tactile sensor is mainly analyzed, compared to the unimodal (EIT-only) case.

        First, the contact localization performance is evaluated using the testbed and the standard indentation protocol. Figure 5 shows the true and estimated location of each contact. The average localization error was 6.9 mm, which is slightly worse than prior reports \cite{park2021adaptive}. We attribute this degradation primarily to the mechanical properties of the pneumatic pads. Because 3D printed TPU pads are stiffer than previously used materials (e.g., neoprene foam), they distribute the applied force over a larger area, thereby increasing the contact area at the carbon–metal interface. These results suggest that the soft-pad design should be further optimized for EIT sensing. The tendency for localized results to shift from the boundary toward the interior is primarily driven by the nonuniform current density in EIT, which causes high sensitivity gradients near the electrodes \cite{park2022Neural}.

        Next, we evaluated the accuracy of the force estimation. The contact force is calculated as the spatial sum of the reconstructed conductivity image. The raw data obtained from the grid indentation protocol $15 \times 15$ was used in both the EIT-only case and the hybrid EIT-pneumatic hybrid case. Figure 6 shows the estimated contact force, plotting a coefficient of variation (CV) for both cases. The EIT-only case showed 0.31, while the proposed hybrid case showed 0.14. The EIT-only case exhibits a larger sensitivity difference at specific sites, mainly due to variance in position-dependent piezoresistive characteristics. However, the hybrid case shows consistent reconstruction sensitivity. The spots in which irregular sensitivity is observed in the EIT-only case are effectively compensated owing to the pneumatic sensor.

    \section{DEMONSTRATION} 

    \begin{figure}[tb!]
        \centering
        \includegraphics[width=3.3in]{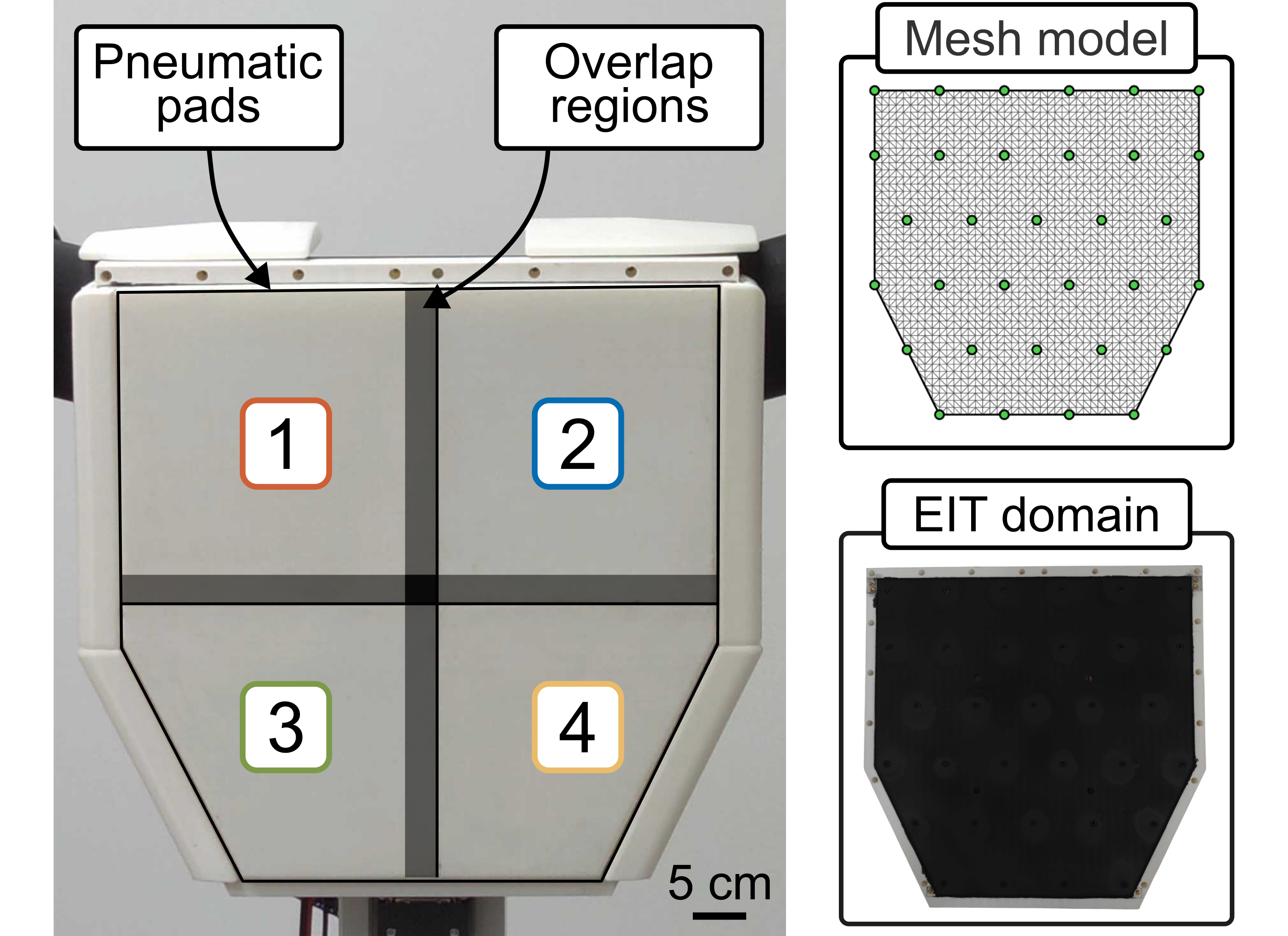}
        \caption{Humanoid chest with EIT–pneumatic skin. The system employs four pneumatic pads (indices annotated in the photo). The carbon-coated EIT surface and the corresponding finite-element mesh are also presented.}
        \label{figure label 07}
    \end{figure}
        
    \begin{figure*}[htbp]
            \centering
            \includegraphics[width=6.9in]{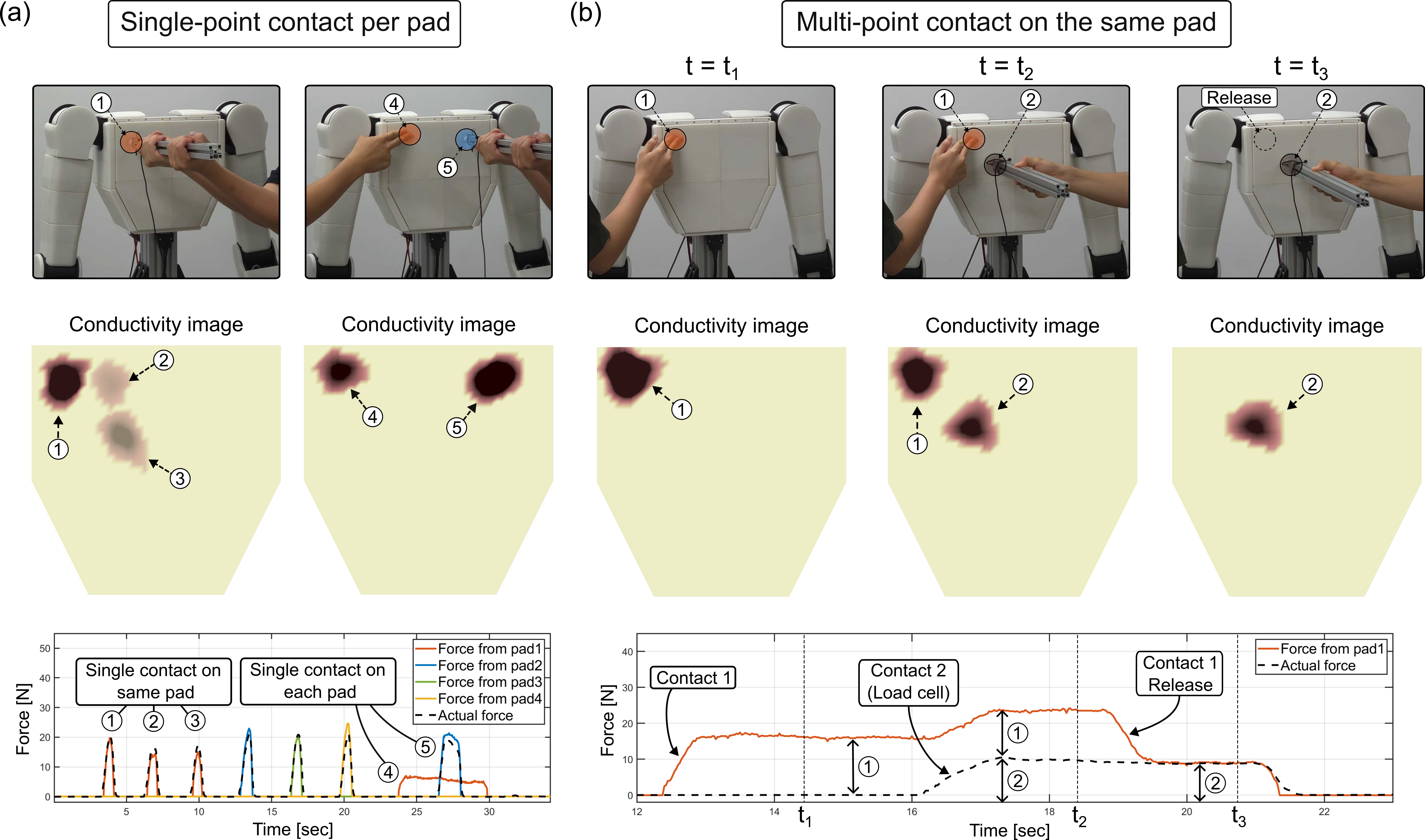}
            \caption{Demonstration on skin-integrated robot. (a) single-point contact per pad. EIT-only conductivity images show position-dependent magnitude discrepancies, while the time-series plot shows the predicted force (black dashed) matching per-pad estimates with no cross-pad interference; (b) multi-point contact on a same pad. The pneumatic signals is well approximated as the sum of two sequential contacts.}
            \label{figure label 08}
        \end{figure*}   

    From a practical standpoint, systems integration is a key consideration for robotic-skin design. Achieving neat and uniform coverage over large areas is challenging, especially on complex curved geometries. Accordingly, our approach is well suited for whole-body skin, as all components can be fabricated using only 3D printing or spray-coating. Moreover, the design lowers the barrier to adoption by relying on cost-effective, off-the-shelf materials. 
    
    For the sensing electronics, the EIT stage requires only a simple electrode-switching circuit, whereas the pneumatic channel uses off-the-shelf pressure sensors. Because few electrodes and sensors are required, the wiring and tubing are minimal. Consequently, the electronics are simple, compact, and cost-effective, enabling straightforward integration into the robot.
   
    \subsection{Skin Integration on the Humanoid Chest}
        To support the above claims, the proposed EIT–pneumatic skin was deployed on the robot’s chest (see Fig.~\ref{figure label 07}). The system drives EIT with 32 electrodes while concurrently acquiring pressure signals from four soft pneumatic pads.
        The fabrication process is identical to those described earlier, and the completed skin has dimensions 280 mm $\times$ 280 mm.

        For EIT tactile sensing, we constructed a new forward model (finite element mesh) and computed the corresponding reconstruction matrix. Measurements were acquired using an FPGA-based EIT drive circuit and transmitted to the host PC via TCP/IP. The resulting images were then reconstructed in real-time at a 100 Hz frame rate by applying the precomputed matrix to the measurements. Finally, the images were lightly post-processed (e.g., up-scaling, smoothing) for visualization. 
        
        The pressure signals from four pneumatic pads were acquired by a microcontroller and transmitted to the host PC via serial communication. Then, these data were synchronized with the EIT stream and logged together.

    \subsection{Tactile Sensing under Various Contact Scenarios}
        We evaluated the EIT-pneumatic hybrid skin in two contact scenarios. The first scenario is single-point contact per pad (each pad was pressed at one location at a time). Another scenario is when multi-point contact on the same pad.
        
        \vspace{6pt}
        \noindent (1) Single-point contact per pad
        \vspace{2pt}\\
        As shown in Fig. 8(a), each pad was indented with a rod equipped with a load cell. The middle panel shows an EIT-only reconstruction (not yet corrected by pneumatic signals). Despite similar applied forces, the reconstructed magnitudes differ markedly, revealing position dependent sensitivity in the EIT skin. By contrast, the bottom time series plot shows that the predicted force (black dashed line) closely matches the estimates derived from each pad, and this correspondence was consistent across locations within a pad. We also observed no interference between pads, indicating that pneumatic-based force estimation operates independently for each pad.

        \vspace{6pt}
        \noindent (2) Multi-point contact on the same pad
        \vspace{2pt}\\
        In this scenario, we created and then released two contacts in sequence on the same pad, as shown in Fig.~8(b). A quasi-static finger contact was applied first (contact 1). Next, we pressed a different location on the same pad (contact 2), which produced a corresponding increase in the pneumatic signal. We then released contact 1, followed by contact 2. After releasing contact 1, the force associated with contact 2 was still accurately estimated from the pneumatic signal. These results indicate that the pneumatic signal can be well approximated as the sum of two contacts.

    \section{DISCUSSION}
    
    
        
        
        


    The proposed EIT–pneumatic hybrid skin provides a practical path toward whole-body tactile sensing. All components are produced by 3D printing or spray-coating, and the hardware remains simple and compact. By fusing complementary cues, the system reduces errors in force estimation due to sensitivity irregularities, without heavy reliance on complex ML models or large calibration datasets \cite{lee2022predicting, chen2021location}. Adding the pneumatic modality does not increase fabrication or structural complexity, as demonstrated by the prototype in Fig.~\ref{figure label 07}.

    One limitation of the pneumatic signal is a drop in sensitivity near overlap regions. Since the current method relies mainly on pneumatic signals, errors in those areas can bias the force estimation. Nevertheless, this issue can be addressed by using spatial cues from the EIT reconstructions; for example, we can adjust the pressure-force mapping according to the contact location and size derived from EIT reconstruction. With this location-aware calibration, we expect the pneumatic and EIT signals complement to each other and improve overall accuracy. The same mechanism also lets the pads serve as an in situ calibrator: instead of collecting large datasets offline, the system can update its calibration during operation, enabling online continual learning of tactile inference. 
    
    This study suggests several promising directions for future work. Since the pneumatic signal detects contact faster and more reliably than EIT, it can serve as a contact event observer that triggers adaptive, event--driven EIT measurement patterns based on contact presence and location. In addition, it would be possible to improve accuracy and robustness of tactile sensing by training end-to-end sensor fusion model through multiphysics simulation and online continual learning. 

    \section*{ACKNOWLEDGMENT}
    This work was supported in part by the National Research Foundation of Korea (NRF) grant funded by the Korea government (MSIT) (RS-2024-00352818), in part by the NRF grant funded by the MSIT (RS-2025-25448259), in part by Basic Science Research Program through the NRF funded by the Ministry of Education (RS-2025-25420118), and in part by the Institute of Information \& Communications Technology Planning \& Evaluation (IITP) grant funded by the Korea government (MSIT) (RS-2025-25442149, LG AI STAR Talent Development Program for Leading Large-Scale Generative AI Models in the Physical AI Domain).

    \bibliographystyle{IEEEtran}
    \bibliography{reference.bib}

@INPROCEEDINGS{park2011designing,
      author={Park, Jung-Jun and Haddadin, Sami and Song, Jae-Bok and Albu-Schäffer, Alin},
      booktitle={2011 IEEE International Conference on Robotics and Automation}, 
      title={Designing optimally safe robot surface properties for minimizing the stress characteristics of human-robot collisions}, 
      year={2011},
      volume={},
      number={},
      pages={5413-5420},
      doi={10.1109/ICRA.2011.5980282}}

@InProceedings{Haddadin2011Safe,
      author={Haddadin, Sami and Albu-Sch{\"a}ffer, Alin and Hirzinger, Gerd},
      editor={Kaneko, Makoto and Nakamura, Yoshihiko},
      title={Safe Physical Human-Robot Interaction: Measurements, Analysis and New Insights},
      booktitle={Robotics Research},
      year={2011},
      publisher={Springer Berlin Heidelberg},
      address={Berlin, Heidelberg},
      pages={395--407},
      isbn={978-3-642-14743-2}}

@ARTICLE{svarny2022effect,
      title = {Effect of active and passive protective soft skins on collision forces in human–robot collaboration},
      journal = {Robotics and Computer-Integrated Manufacturing},
      volume = {78},
      pages = {102363},
      year = {2022},
      issn = {0736-5845},
      doi = {https://doi.org/10.1016/j.rcim.2022.102363},
      author = {Petr Svarny and Jakub Rozlivek and Lukas Rustler and Martin Sramek and Özgür Deli and Michael Zillich and Matej Hoffmann}}

@ARTICLE{adler2006eidors,
      doi = {10.1088/0967-3334/27/5/S03},
      url = {https://dx.doi.org/10.1088/0967-3334/27/5/S03},
      year = {2006},
      month = {apr},
      publisher = {},
      volume = {27},
      number = {5},
      pages = {S25},
      author = {Adler, Andy and Lionheart, William R B},
      title = {Uses and abuses of EIDORS: an extensible software base for EIT},
      journal = {Physiological Measurement}}

@ARTICLE{tawil2015electrical,
      author={Silvera-Tawil, David and Rye, David and Soleimani, Manuchehr and Velonaki, Mari},
      journal={IEEE Sensors Journal}, 
      title={Electrical Impedance Tomography for Artificial Sensitive Robotic Skin: A Review}, 
      year={2015},
      volume={15},
      number={4},
      pages={2001-2016},
      doi={10.1109/JSEN.2014.2375346}}

@article{cui2023recent,
      title={Recent developments in impedance-based tactile sensors: A review},
      author={Cui, Ziqiang and Yu, Yongkang and Wang, Huaxiang},
      journal={IEEE Sensors Journal},
      volume={24},
      number={3},
      pages={2350--2366},
      year={2023},
      publisher={IEEE}
    }

@article{liu2020artificial,
      title={Artificial sensitive skin for robotics based on electrical impedance tomography},
      author={Liu, Kai and Wu, Yang and Wang, Song and Wang, Huan and Chen, Huaijin and Chen, Bai and Yao, Jiafeng},
      journal={Advanced Intelligent Systems},
      volume={2},
      number={4},
      pages={1900161},
      year={2020},
      publisher={Wiley Online Library}
    }

@ARTICLE{park2022biomimetic,
      author = {K. Park  and H. Yuk  and M. Yang  and J. Cho  and H. Lee  and J. Kim },
      title = {A biomimetic elastomeric robot skin using electrical impedance and acoustic tomography for tactile sensing},
      journal = {Science Robotics},
      volume = {7},
      number = {67},
      pages = {eabm7187},
      year = {2022},
      doi = {10.1126/scirobotics.abm7187}
    }

@ARTICLE{park2022Neural,
      author={Park, Kyungseo and Kim, Jung},
      journal={IEEE Transactions on Robotics}, 
      title={Neural-Gas Network-Based Optimal Design Method for ERT-Based Whole-Body Robotic Skin}, 
      year={2022},
      volume={38},
      number={6},
      pages={3463-3478},
      doi={10.1109/TRO.2022.3186806}}

@article{park2021adaptive,
      title={Adaptive optimal measurement algorithm for ERT-based large-area tactile sensors},
      author={Park, Kyungseo and Lee, Hyosang and Kuchenbecker, Katherine J and Kim, Jung},
      journal={IEEE/ASME Transactions on Mechatronics},
      volume={27},
      number={1},
      pages={304--314},
      year={2021},
      publisher={IEEE}
    }

@ARTICLE{tawil2011improved,
      author={Tawil, David Silvera and Rye, David and Velonaki, Mari},
      journal={IEEE Transactions on Robotics}, 
      title={Improved Image Reconstruction for an EIT-Based Sensitive Skin With Multiple Internal Electrodes}, 
      year={2011},
      volume={27},
      number={3},
      pages={425-435},
      doi={10.1109/TRO.2011.2125310}}

@INPROCEEDINGS{nagakubo2007deformable,
      author={Nagakubo, Akihiko and Alirezaei, Hassan and Kuniyoshi, Yasuo},
      booktitle={2007 IEEE International Conference on Robotics and Biomimetics (ROBIO)}, 
      title={A deformable and deformation sensitive tactile distribution sensor}, 
      year={2007},
      volume={},
      number={},
      pages={1301-1308},
      doi={10.1109/ROBIO.2007.4522352}}

@article{kim2024extremely,
    author = {Kyubeen Kim  and Jung-Hoon Hong  and Kyubin Bae  and Kyounghun Lee  and Doohyun J. Lee  and Junsu Park  and Haozhe Zhang  and Mingyu Sang  and Jeong Eun Ju  and Young Uk Cho  and Kyowon Kang  and Wonkeun Park  and Suah Jung  and Jung Woo Lee  and Baoxing Xu  and Jongbaeg Kim  and Ki Jun Yu },
    title = {Extremely durable electrical impedance tomography–based soft and ultrathin wearable e-skin for three-dimensional tactile interfaces},
    journal = {Science Advances},
    volume = {10},
    number = {38},
    pages = {eadr1099},
    year = {2024},
    doi = {10.1126/sciadv.adr1099}
    }

@ARTICLE{vauhkonen1998tikhonov,
      author={Vauhkonen, M. and Vadasz, D. and Karjalainen, P.A. and Somersalo, E. and Kaipio, J.P.},
      journal={IEEE Transactions on Medical Imaging}, 
      title={Tikhonov regularization and prior information in electrical impedance tomography}, 
      year={1998},
      volume={17},
      number={2},
      pages={285-293},
      doi={10.1109/42.700740}}

@ARTICLE{borsic2010invivo,
      author={Borsic, Andrea and Graham, Brad M. and Adler, Andy and Lionheart, William R. B.},
      journal={IEEE Transactions on Medical Imaging}, 
      title={In Vivo Impedance Imaging With Total Variation Regularization}, 
      year={2010},
      volume={29},
      number={1},
      pages={44-54},
      doi={10.1109/TMI.2009.2022540}}

@ARTICLE{liu2018limage,
      author={Liu, Shengheng and Jia, Jiabin and Zhang, Yimin D. and Yang, Yunjie},
      journal={IEEE Transactions on Medical Imaging}, 
      title={Image Reconstruction in Electrical Impedance Tomography Based on Structure-Aware Sparse Bayesian Learning}, 
      year={2018},
      volume={37},
      number={9},
      pages={2090-2102},
      doi={10.1109/TMI.2018.2816739}}

@INPROCEEDINGS{park2020ERT,
      author={Park, Kyungseo and Park, Hyunkyu and Lee, Hyosang and Park, Sungbin and Kim, Jung},
      booktitle={2020 IEEE International Conference on Robotics and Automation (ICRA)}, 
      title={An ERT-based Robotic Skin with Sparsely Distributed Electrodes: Structure, Fabrication, and DNN-based Signal Processing}, 
      year={2020},
      volume={},
      number={},
      pages={1617-1624},
      doi={10.1109/ICRA40945.2020.9197361}}

@ARTICLE{park2021Deep,
      author={Park, Hyunkyu and Park, Kyungseo and Mo, Sangwoo and Kim, Jung},
      journal={IEEE Transactions on Robotics}, 
      title={Deep Neural Network Based Electrical Impedance Tomographic Sensing Methodology for Large-Area Robotic Tactile Sensing}, 
      year={2021},
      volume={37},
      number={5},
      pages={1570-1583},
      doi={10.1109/TRO.2021.3060342}}

@article{lee2022predicting,
      title={Predicting the force map of an ERT-based tactile sensor using simulation and deep networks},
      author={Lee, Hyosang and Sun, Huanbo and Park, Hyunkyu and Serhat, Gokhan and Javot, Bernard and Martius, Georg and Kuchenbecker, Katherine J},
      journal={IEEE Transactions on Automation Science and Engineering},
      volume={20},
      number={1},
      pages={425--439},
      year={2022},
      publisher={IEEE}
    }

@INPROCEEDINGS{lee2020calibrating,
      author={Lee, Hyosang and Park, Hyunkyu and Serhat, Gokhan and Sun, Huanbo and Kuchenbecker, Katherine J.},
      booktitle={2020 IEEE International Conference on Robotics and Automation (ICRA)}, 
      title={Calibrating a Soft ERT-Based Tactile Sensor with a Multiphysics Model and Sim-to-real Transfer Learning}, 
      year={2020},
      volume={},
      number={},
      pages={1632-1638},
      doi={10.1109/ICRA40945.2020.9196732}
    }

@ARTICLE{dong2025learning,
      author={Dong, Huazhi and Wu, Xiaopeng and Hu, Delin and Liu, Zhe and Giorgio-Serchi, Francesco and Yang, Yunjie},
      journal={IEEE Transactions on Instrumentation and Measurement},
      title={Learning-Enhanced Electronic Skin for Tactile Sensing on Deformable Surface Based on Electrical Impedance Tomography},
      year={2025},
      volume={74},
      number={},
      pages={1-9},
      doi={10.1109/TIM.2025.3546404}}

@ARTICLE{zheng2025large,
      author={Zheng, Wendong and Guo, Di and Yang, Wuqiang and Liu, Huaping},
      journal={IEEE/ASME Transactions on Mechatronics},
      title={A Large-Area Robotic Skin for Intelligent Tactile Interaction of Collaborative Robots},
      year={2025},
      volume={},
      number={},
      pages={1-11},
      doi={10.1109/TMECH.2024.3520953}}

@article{li2019novel,
      title={A novel deep neural network method for electrical impedance tomography},
      author={Li, Xiuyan and Zhou, Yong and Wang, Jianming and Wang, Qi and Lu, Yang and Duan, Xiaojie and Sun, Yukuan and Zhang, Jingwan and Liu, Zongyu},
      journal={Transactions of the Institute of Measurement and Control},
      volume={41},
      number={14},
      pages={4035--4049},
      year={2019},
      publisher={SAGE Publications Sage UK: London, England}
    }

@article{hrabuska2018image,
    title = {Image Reconstruction for Electrical Impedance Tomography: Experimental Comparison of Radial Basis Neural Network and Gauss – Newton Method},
    journal = {IFAC-PapersOnLine},
    volume = {51},
    number = {6},
    pages = {438-443},
    year = {2018},
    note = {15th IFAC Conference on Programmable Devices and Embedded Systems PDeS 2018},
    issn = {2405-8963},
    doi = {https://doi.org/10.1016/j.ifacol.2018.07.114},
    author = {Radek Hrabuska and Michal Prauzek and Marketa Venclikova and Jaromir Konecny}
    }

@article{duan2019artificial,
      title={Artificial skin through super-sensing method and electrical impedance data from conductive fabric with aid of deep learning},
      author={Duan, Xi and Taurand, Sebastien and Soleimani, Manuchehr},
      journal={Scientific reports},
      volume={9},
      number={1},
      pages={8831},
      year={2019},
      publisher={Nature Publishing Group UK London}
    }

@INPROCEEDINGS{Kim20153D,
      author={Kim, Joohyung and Alspach, Alexander and Yamane, Katsu},
      booktitle={2015 IEEE/RSJ International Conference on Intelligent Robots and Systems (IROS)}, 
      title={3D printed soft skin for safe human-robot interaction}, 
      year={2015},
      volume={},
      number={},
      pages={2419-2425},
      doi={10.1109/IROS.2015.7353705}}

@INPROCEEDINGS{goncalves2022punyo,
      author={Goncalves, Aimee and Kuppuswamy, Naveen and Beaulieu, Andrew and Uttamchandani, Avinash and Tsui, Katherine M. and Alspach, Alex},
      booktitle={2022 IEEE 5th International Conference on Soft Robotics (RoboSoft)}, 
      title={Punyo-1: Soft tactile-sensing upper-body robot for large object manipulation and physical human interaction}, 
      year={2022},
      volume={},
      number={},
      pages={844-851},
      doi={10.1109/RoboSoft54090.2022.9762117}}

@INPROCEEDINGS{rustler2024adaptive,
      author={Rustler, Lukas and Misar, Matej and Hoffmann, Matej},
      booktitle={2024 IEEE-RAS 23rd International Conference on Humanoid Robots (Humanoids)}, 
      title={Adaptive Electronic Skin Sensitivity for Safe Human-Robot Interaction}, 
      year={2024},
      volume={},
      number={},
      pages={475-482},
      doi={10.1109/Humanoids58906.2024.10769602}}

@ARTICLE{park2024lowcost,
      author={Park, Kyungseo and Shin, Kazuki and Yamsani, Sankalp and Gim, Kevin and Kim, Joohyung},
      journal={IEEE Transactions on Robotics}, 
      title={Low-Cost and Easy-to-Build Soft Robotic Skin for Safe and Contact-Rich Human–Robot Collaboration}, 
      year={2024},
      volume={40},
      number={},
      pages={2327-2338},
      doi={10.1109/TRO.2024.3378174}}

@INPROCEEDINGS{park2024fully,
      author={Taylor, Sean and Park, Kyungseo and Yamsani, Sankalp and Kim, Joohyung},
      booktitle={2024 IEEE International Conference on Robotics and Automation (ICRA)}, 
      title={Fully 3D printable Robot Hand and Soft Tactile Sensor based on Air-pressure and Capacitive Proximity Sensing}, 
      year={2024},
      volume={},
      number={},
      pages={18100-18105},
      doi={10.1109/ICRA57147.2024.10610731}}

@article{chen2025large,
      title={Large-area Tomographic Tactile Skin with Air Pressure Sensing for Improved Force Estimation},
      author={Chen, Haofeng and Himmel, Bedrich and Kubik, Jiri and Hoffmann, Matej and Lee, Hyosang},
      journal={arXiv preprint arXiv:2503.13036},
      year={2025}
    }

@article{chen2021location,
      title={Location-dependent performance of large-area piezoresistive tactile sensors based on electrical impedance tomography},
      author={Chen, Ying and Liu, Haibin},
      journal={IEEE Sensors Journal},
      volume={21},
      number={19},
      pages={21622--21630},
      year={2021},
      publisher={IEEE}
    }

@ARTICLE{yang2025body,
      author={Yang, Min Jin and Chung, Hyunjo and Kim, Yoonjin and Park, Kyungseo and Kim, Jung},
      journal={IEEE Transactions on Robotics}, 
      title={A Body-Scale Robotic Skin Using Distributed Multimodal Sensing Modules: Design, Evaluation, and Application}, 
      year={2025},
      volume={41},
      number={},
      pages={96-109},
      doi={10.1109/TRO.2024.3502204}}

@ARTICLE{costa2023variable,
      title={Variable sensitivity multimaterial robotic e-skin combining electronic and ionic conductivity using electrical impedance tomography},
      author={Costa Cornell{\`a}, Aleix and Hardman, David and Costi, Leone and Brancart, Joost and Van Assche, Guy and Iida, Fumiya},
      journal={Scientific Reports},
      volume={13},
      number={1},
      pages={20004},
      year={2023},
      publisher={Nature Publishing Group UK London}
    }

@ARTICLE{chen2023novel,
      author={Chen, Huaijin and Langlois, Kevin and Brancart, Joost and Roels, Ellen and Verstraten, Tom and Vanderborght, Bram},
      journal={IEEE Sensors Journal},
      title={A Novel Physical Human–Robot Interface With Pressure Distribution Measurement Based on Electrical Impedance Tomography},
      year={2023},
      volume={23},
      number={18},
      pages={21914-21923},
      doi={10.1109/JSEN.2023.3303226}}

\end{document}